\documentclass{esannV2}
\usepackage[dvips, pdftex]{graphicx}
\usepackage[latin1]{inputenc}
\usepackage{amssymb,amsmath,array}
\usepackage{mathrsfs}
\usepackage{booktabs}
\usepackage{wrapfig}

\usepackage[inkscapelatex=false]{svg}

\usepackage{titlesec}
\titlespacing*{\section}{0pt}{1.2ex plus .2ex minus .1ex}{0.8ex plus .1ex}
\titlespacing*{\subsection}{0pt}{1.0ex plus .2ex minus .1ex}{0.6ex plus .1ex}

\usepackage{hyperref}
\usepackage{xfrac}

\usepackage[
  backend=biber,
  style=numeric-comp,    
  sorting=none,          
  maxbibnames=3,         
  minbibnames=1,
  maxcitenames=2,        
  natbib=true,
  giveninits=true
]{biblatex}
\AtBeginBibliography{
  \small
  \setlength{\itemsep}{0.3ex} 
  \setlength{\parskip}{0pt}   
}

\renewcommand{\vec}[1]{\boldsymbol{#1}}
\newcommand{\R}{{\rm I\!R}}

\begin{document}

%style file for ESANN manuscripts
\title{
%Hub-Enhanced Locally Aligned Ant Technique for Efficient Manifold Detection \\
Hub-Aware Hybrid Search: Accelerating the Locally Aligned Ant Technique \\
%Hybrid Likelihood-Driven Locally Aligned Ant Technique for Robust Structure Detection \\
%Hybrid Probabilistic Guidance: A Fast Locally Aligned Ant Technique for Structure Detection
}

%***********************************************************************
% AUTHORS INFORMATION AREA
%***********************************************************************
\author{Simone Vilardi$^1$, Reynier Peletier$^2$, Felipe Contreras$^{1,3}$ and Kerstin Bunte$^1$
%
% Optional short acknowledgment: remove next line if non-needed
%\thanks{This is an optional funding source acknowledgement.}
%
% DO NOT MODIFY THE FOLLOWING '\vspace' ARGUMENT
\vspace{.3cm}\\
%
% Addresses and institutions (remove "1- " in case of a single institution)
1- University of Groningen - Bernoulli Institute \\
for Mathematics, Computer Science and Artificial Intelligence
% \\% Broerstraat 5, 9712 CP Groningen - Netherlands
%
% Remove the next three lines in case of a single institution
\vspace{.1cm}\\
2- University of Groningen - Kapteyn Astronomical Institute %\\Broerstraat 5, 9712 CP Groningen - Netherlands
\vspace{.1cm}\\
3-  Instituto de F\'isica y Astronom\'ia, Universidad de Valpara\'iso %, casilla 5030, %\\Avda.~Gran Breta\~na 1111, Playa Ancha, Valpara\'iso 2360102, Chile
}
%***********************************************************************
% END OF AUTHORS INFORMATION AREA
%***********************************************************************

\maketitle

\begin{abstract}
Finding manifold structures %embedded with
in noisy and high-dimensional point clouds is a challenging but important problem. In astronomical observation survey and simulation data the detection of filaments, streams (1D), walls (2D) and clusters (3D) gives rise to deeper understanding of the evolution of our universe. 
% Finding filaments, streams and clusters within the cosmic web is crucial, yet challenging, as these structures are embedded in noisy, high-dimensional survey and simulation data. 
The Locally Aligned Ant Technique (LAAT) uses biologically inspired agents to efficiently recover faint and multidimensional structures. However, very dense hubs (e.g. nodes or globular clusters) dominate the ants' activity, creating unnecessary computational overheads. %bottlenecks. 
In this paper we propose a two-stage solution.
First a fast preprocessing step locates the hubs and replaces them with a tailored likelihood model. 
% We address this issue with a two-stage solution: first, a fast preprocessing step locates the hubs and replaces them with a tailored likelihood model; then, 
Subsequently, a mixed likelihood-pheromone %scheme
strategy guides the ants to efficiently bridge the dense regions. 
We demonstrate improvements in detection efficiency and robustness % This reduces overheads and improves the robustness and detection speed 
of LAAT with synthetic and a large-scale astronomical N-body simulation of the cosmic web. %large-scale structures.
\end{abstract}

\section{Introduction}

%To understand the large-scale structure of the universe it is paramount to identify filaments, streams, and clusters within the cosmic web. 
%Dark matter and dark energy influence the formation of galaxies and the distribution of intergalactic gas, creating filamentary matter bridges that reveal the underlying dark matter distribution and provide information on cosmological parameters. 
%However, these features are difficult to detect because they are embedded in very noisy, high-dimensional data from large astronomical surveys and N-body simulations. \citep{Bond1996, Cautun2014, Springel2005, Schaye2015}
To understand the large-scale structure of the Universe, it is paramount to identify filaments, streams, and clusters within the cosmic web, which are difficult to detect due to noise and high dimensionality in astronomical surveys and N-body simulations.\citep{Bond1996, Cautun2014, Springel2005, Schaye2015}
%The 1-DREAM (1D Recovery, Extraction, and Analysis of Manifolds) pipeline was developed to detect and model %isolate % one-dimensional 
%1D manifolds in such settings \citep{Taghribi2022,CANDUCCI2022100658}. %, 2024raj}. %The pipeline's modules, including LAAT (Locally Aligned Ant Technique), EM3 A, DimIndex, 1D Multi-Manifold Crawling, and SGTM, emphasise high-contrast regions and efficient background filtering. %This makes the toolkit ideal for identifying filamentary and stream-like structures.
% LAAT, which is the first of the pipeline's modules and is based on ant colony optimisation ideas, reduces the background noise by running many stochastic 'ants' that favour jumps aligned with the local principal directions and with previously visited (pheromone-rich) points \citep{Taghribi2022}. 
%The Locally Aligned Ant Technique (LAAT) is the first module of the pipeline to detect and extract likely members of manifold structures embedded in large amounts of background noise. 
%It is an evolutionary computation algorithm based on the idea of ant colony optimization, with transition probabilities using local neighbourhood geometry and a reinforcing pheromone component deposited on visitation \citep{Taghribi2022}.
The 1-DREAM pipeline was developed to detect and model one-dimensional manifolds in such settings \citep{Taghribi2022, CANDUCCI2022100658}, with LAAT as its first module for identifying likely manifold members via an ant-colony optimisation scheme combining local geometry and pheromone reinforcement \cite{Taghribi2022}.
%Practical hyperparameters include the number of ants, the number of epochs, the number of steps per ant, initialisation among denser neighbourhoods, and the pheromone increment $\gamma$. By thresholding low-pheromone points, LAAT can reveal filamentary features even in noisy data.
However, real datasets often contain very dense hubs where ants %quickly concentrate, such that %; 
rapidly accumulate, causing these clusters to dominate pheromone accumulation and subsequent thresholding. 
This wastes computation resources and hampers %masks 
the desired one-dimensional manifold recovery. 
Recent extensions proposing a dynamic radius \citep{felipe23, felipe25} improved the noise handling and user-friendliness of LAAT, but the ants still spent a significant amount of steps (time) in dense regions. 
% Despite recent improvements in noise handling and the introduction of a %the new dynamic radius approach \citep{felipe23, felipe25}, this pheromone overconcentration remains a significant issue.
We present a revised LAAT that mitigates hub dominance\footnote{code publicly available at \url{https://git.lwp.rug.nl/cs.projects/Hub-LAAT}} by: % Our approach involves the following steps: 
(1) identifying dense regions, 
(2) fitting a likelihood model to each region, 
(3) running ants on the combined likelihood-point cloud domain, and 
(4) introducing a modified transition probability incorporating those models %using those models 
to guide the ants towards fainter structures. 
These modifications retain the alignment-and-pheromone mechanism of LAAT, while avoiding %discouraging 
the repeated exploration %exploitation 
of dense hubs. This improves the recovery of filamentary manifolds in challenging cosmological datasets.

\section{Methodology}
LAAT guides stochastic agents across a dataset using pheromone deposition and %-based 
evaporation schemes inspired by biology.
Given a dataset $\mathcal{X}=\{\vec{x}_i\}_{i=1}^N%,\dots,\vec{x}_n\}
$, for each point, $\vec{x}_i\in\R^D$ we define a spherical neighbourhood $\mathcal{N}_r^{(i)}$ of radius $r$ and compute %$d$ 
local %principal components 
eigen-vectors and -values $\{(\vec{v}_d,\lambda)_d)\}_{d=1}^D$ within it. % within that neighbourhood. 
LAAT drives ants with two preferences: 
(i) Alignment of $(\vec{x}_j - \vec{x}_i)$ with $\vec{v}_d$ represented by angle $|\cos\alpha_d^{(i,j)}|$ and weighted by %corresponding 
$\lambda_d$, thus 
$A^{(i,j)}=%\displaystyle
\sum_{d=1}^D\frac{|\cos\alpha_d^{(i,j)}|}{\sum_{d\prime}|\cos\alpha_{d\prime}^{(i,j)}|} \cdot 
\frac{\lambda_d}{\sum_{d\prime}\lambda_{d\prime}}$, and neighborhood relative $\bar{A}^{(i,j)} = \frac{A^{(i,j)}}{\sum_{j\prime\in\mathcal{N}^{(i)}_r} A^{(i,j\prime)}}$.
% dominant local eigenvector(s) (measured from the angle between the path and the local orientation) and 
(ii) And relative accumulated pheromone $\bar{F}^j(t)$ at time $t$ at point $\vec{x}_j\in \mathcal{N}^{(i)}_r$, that can evaporate on rarely visited points (see \cite{Taghribi2022}).
Both combined result in a movement preference from $\vec{x}_i$ to $\vec{x}_j$ of an ant at time $t$ with contributions controlled by
% (with an evaporation factor to downweight rarely visited points). The combined preference between points $i$ and $j$ at time $t$ is written $\mathcal{V}^{(i,j)}(t)$, where a parameter 
$\kappa \in[0,1]$: %controls the relative weight of the pheromone $F^{j}(t)$ and alignment $A^{(i,j)}(t)$ contributions:
\begin{equation}
    V^{(i,j)}(t) = (1-\kappa)\bar{F}^{j}(t) + \kappa \bar{A}^{(i,j)}. %(t).
    \label{eq:Vpref}
\end{equation}
The transition probabilities are viewed as negative ``energies"
% The resulting transition %jump probability from $i$ to $j$ is: % then defined as:
\begin{equation}
    P(j\mid i,t)=\sfrac{\exp%e^
    {\left(\beta\,V^{(i,j)}(t)\right)}}
    {\textstyle\sum_{j'\in\mathcal{N}_r^{(i)}} \exp%e^
    {\left(\beta\,V^{(i,j')}(t)\right)}},
    \label{eq:old_prob}
\end{equation}
% where $\beta$ plays the role of an
with inverse temperature $\beta$. Here $j'$ indexes points in the local neighbourhood $\mathcal{N}_r^{(i)}$ of point $i$.
To allow efficient parallelization with multiple ants, the pheromone is updated in epochs encompassing a number of ant steps. The evolutionary process is linear with the number of points, with a one time squared preprocessing cost including local PCA %. Implementation designs can trade of memory and computational costs 
\citep{Taghribi2022}.
% Although LAAT is efficient, t
The presence of dense hubs induces two issues:
(1) % in dense hubs 
pheromone accumulates, making them local attractors and reducing visitations in fainter structures, and 
(2) structure members are detected by pheromone thresholding, which tends to keep a significant amount of points in these small dense regions, slowing down any subsequent analysis unnecessarily. 
% excessively, so thresholding preserves clusters rather than one dimensional filaments.
% We address this issue by fitting likelihood models to the dense regions %st hubs 
% and allowing ants to explore a mixed space. %, comprising point and likelihood regions.
% This hybrid movement discourages already known hub exploration %exploitation 
% and improves filament recovery.

\subsection{Hub identification and modelling}
% Our approach prioritises a fast, lightweight pipeline for identifying dense regions with minimal memory and CPU overhead. Our first objective is to assign a score to each point indicating the likelihood that it belongs to a dense hub. To obtain this score, we run a Markov chain guided by LAAT's alignment term $A_{ij}$. For a pair of points $i,j$, we set $w_{ij} = e^{-\delta A_{ij}}$, with $\delta = 10$, and we store the resulting triplets $(i,j,w_{ij})$ in a sparse matrix. The stationary distribution (dominant eigenvector) of this matrix, computed via Power Iteration, yields the score map highlighting dense candidates.
% We reuse the thresholding heuristic from \citet{felipe23} (originally used for pheromone thresholding) to decide which scores qualify as dense, applying it early in the pipeline so that it adapts to a wide range of dataset scenarios. Finally, we cluster high-scoring points using friends-of-friends clustering, where the linking length can be either the original LAAT radius or the cosmological choice $l = b\left(\frac{V}{N}\right)^{1/d}$,  where $V$ is the data volume (or surface), $N$ the number of points, $d$ the dimensionality and $b$ is a free parameter. If the user sets $b$ to $0$, the linking length $l$ is set as the neighbourhood radius $r$.
%We note that $\psi$, $\eta$ and $b$ may require user tuning; these are the only new free parameters introduced by this stage.
Our approach is designed for efficiency, identifying dense regions to reduce computational burden in the subsequent evolutionary computation. %with minimal computational overhead. 
In the special case of $\kappa=1$ in \eqref{eq:Vpref} the ants ignore the pheromone and transition probabilities are independent of $t$ and stay constant with $P_{ij}=P(j|i)$. It can be considered a Markov Chain (MC) and studied as stationary distribution of the point cloud (several if isolated). It is independent of the starting point and will converge to a steady-state vector $\vec{\pi}$, the dominant left eigenvector of the transition matrix containing the visitation frequency of each data point. 
We compute this eigenvector using the Power method and use the thresholding strategy proposed in \cite{felipe23} %\citet{felipe23} 
to obtain high visitation values indicating dense regions. 
% Each data point is assigned a score indicating its likelihood of belonging to a dense hub. 
% To obtain this score, we run a Markov chain guided by LAAT's alignment term $A^{(i,j)}$ and define pairwise weights $w_{ij}=e^{\beta A_{ij}}$ for all neighbours $j \in \mathcal{N}(i)$, and $w_{ij}=0$ otherwise. The parameter $\beta>0$ controls how strongly alignment differences are amplified.
% The triplets $(i,j,w_{ij})$ form a sparse matrix $W$ with entries $W_{ij}=w_{ij}$. Starting from an initial vector $\mathbf{q}^{(0)}$ with all components set to $1$, we apply Power Iteration,
% \begin{equation}
%     \mathbf{q}^{(k+1)} = \frac{W\, \mathbf{q}^{(k)}}{\bigl\|W\, \mathbf{q}^{(k)}\bigr\|_2},
% \end{equation}
% until convergence, or
% \begin{equation}
%     \bigl\|\mathbf{q}^{(k+1)} - \mathbf{q}^{(k)}\bigr\|_2 < \varepsilon
% \end{equation}
% for a small tolerance $\varepsilon$ or a maximum number of iterations.
% The components of the converged vector $\mathbf{q}$ then define the scores $s_i = q_i$, and a score map highlighting dense candidates.
% Scores are thresholded using the heuristic of \citet{felipe23}, which was originally applied to pheromone filtering. 
% Given visitation scores $\{s_i\}_{i=1}^N$ and neighbourhoods $\{\mathcal{N}(i)\}$, we first define the global filter threshold
% \begin{equation}
% T_g = \psi\,\bar{s}, 
% \qquad 
% \bar{s} = \frac{1}{N}\sum_{i=1}^N s_i .
% \end{equation}
Concretely, for every particle $\vec{x}_i$ we sort its neighbours $\vec{x}_j\in\mathcal{N}_r^{(i)}$ in ascending order of visitation score $s_j=\pi_j$, obtain the minimum (maximum) value $s_{\min}$ ($s_{\max}$), and 
% For each particle $i$, we collect the neighbour scores $\mathcal{S}_i$, sort them in ascending order, and 
%fit a %natural 
smooth the distribution by fitting a cubic spline $f(s)$. % to the sorted values.
The first inflection point $s_0$ is found by differentiation and marks a steep change.
A parameter $\eta$ defines a local threshold $T_i$, more aggressive or conservative: %, to classify dense region members: % over the domain
% Sampling the derivative $f'(s)$ densely determines the first local minimum at $s_0$. 
%, with the corresponding threshold $s_0 = f(x_0)$.
% We also define $s_{\min} = \min(\mathcal{S}_i)$ and $s_{\max} = \max(\mathcal{S}_i)$.
% The aggressiveness parameter $\eta$ adjusts the threshold according to
\begin{equation}
T_i =
\begin{cases}
s_0 + (s_{\max} - s_0)\,\eta, & 1\ge \eta \ge 0 , \\
s_0 + (s_0 - s_{\min})\,\eta, & -1 \le \eta < 0 .
\end{cases}
\end{equation}
A minimum threshold is enforced as a fraction of the average visitation 
$T_i \ge \frac{\psi}{N}\sum_{j=1}^N \pi_j$. 
%$T_i \ge T_g$.
% Finally, a point $i$ is classified as dense whenever$s_i > T_i$.
We then cluster high-visitation %scoring 
points via the friends-of-friends algorithm with linking length $l=b(\mathcal{V}/N)^{1/D}$, where $\mathcal{V}$ is the data volume (or surface) with free parameter $b\in[0.1,1]$. %, $N$ the number of points, $d$ the dimensionality, and $b$ a free parameter. 
% Reasonable $b$ values lie in the range $b\in[0.1,1.0]$
% If $b=0$, the linking length defaults to the 
Alternatively $l$ can be chosen as LAAT neighbourhood radius $r$. 
We only retain hubs containing at least $N_{\min}$ points.

After detection we fit a likelihood model to each of the $K$ %detected 
hubs, forming the basis for the new hybrid ant-movement scheme. % that operates across both point clouds and likelihood fields. 
% We model each hub using 
We model the distribution using a Bayesian Gaussian mixture with a Dirichlet process prior% , implemented via the \texttt{BayesianGaussianMixture} class in \texttt{scikit-learn}
\footnote{\texttt{BayesianGaussianMixture} in \url{https://github.com/scikit-learn/scikit-learn}.}.
 % so the effective component count adapts to the data. 
The fitted BGM yields a continuous likelihood field, which we divide into 
%and points are categorised as either 
High-Likelihood Points (HLPs) or Low-Likelihood Points (LLP) to remove the dense central region and keep a transition region to lower density outskirts. 
Although hubs are general, we simplify using nested equal-volume Mahalanobis shells centred at the maximum likelihood point of the $k$'s model $\vec{c}_k$ to approximate likelihood level sets. 
% While doing so, a surrounding shell of LLP is retained to act as a transition region towards the lower-density outskirts. Although the hubs may in general have anisotropic shapes, modelling likelihood level sets exactly is computationally expensive. We therefore approximate likelihood level sets by nested shells in the Mahalanobis metric: we compute the Mahalanobis distance of each point from the maximum likelihood point $x_c$, which is equivalent to working in a coordinate system where level sets become spheres. In this space we construct equal-volume shells by setting the initial radius $R_0$ to the median Mahalanobis distance and defining
% \begin{equation}
%     R_\ell=
%     \begin{cases}
%         (\ell+1)^{1/3}\,R_0, & d=3,\\
%         \sqrt{\ell+1}\,R_0, & d=2
%     \end{cases}
%     \qquad \ell=0,1,2,\dots
% \end{equation}
% which yields spherical shells of equal volume in the transformed space and corresponding ellipsoidal shells in the original coordinates.
Let $N_p$ denote the number of particles enclosed up to %the current 
shell $\ell$ with radius $r_\ell$, %and %let 
$N_{\mathrm{hub}}$ %denote 
the total number of hub particles %belonging to 
% the hub. %current hub. %We then define t
and the enclosed fraction $R = N_p/N_{\mathrm{hub}}$.
For growing shells $\mathcal{S}_\ell = \{\,\vec{x}_j : r_{\ell-1} < \|\vec{x_j} - \vec{c}_k\| \le r_\ell\,\}$,
we compute a stopping probability
% \begin{equation}
$p_{\mathrm{stop}}
    = \sfrac{(e^{\alpha/R}-1)}{(e^{1/R}-1)}$ with
    % \qquad
    $\alpha = \frac{R-0.5}{0.5},
$ %\end{equation}
which increases as a larger fraction of the hub has been enclosed. A random draw $u \sim U(0,1)$ terminates the shell expansion when $u < p_{\mathrm{stop}}$. 
The median likelihood of the final shell %then 
defines the hub %specific 
threshold $T_k$ separating HLP ($>T_k$) from LLP ($\le T_k$). % for hub $k$.
% \begin{equation}
%     \mathrm{LLP}_{k}
%     = 
%     \left\{
%         i \;\middle|\;
%         0 < \mathcal{L}_i^{(k)} \le \mathcal{T}_{k}
%     \right\},
%     \qquad
%     \mathrm{HLP}_{k}
%     =
%     \left\{
%         i \;\middle|\;
%         \mathcal{L}_i^{(k)} > \mathcal{T}_{k}
%     \right\}.
% \end{equation}
Removing the HLP from the data before exploration prevents any future ant visitation, reducing computational time and memory requirements.
% All HLPs are then removed from the dataset before any exploration begins, thereby preventing the ants from being trapped and reducing memory usage, hence improving both computational time and storage requirements.
% and protecting against ants from becoming trapped in dense hubs. This significantly reduces both computational time and storage requirements while improving exploration efficiency.
\subsection{New ant transition probabilities} %movement}
All remaining data points $\vec{x}_i$ carry likelihood information for the $K$ modelled hubs $\mathcal{L}_k^{(i)}$ with $k=1\dots K$. 
The friends-of-friends clustering provides a crisp assignment so each point has at most one likelihood value (extension to fuzzy memberships are possible). 
% After the previous step, each point $i$ carries auxiliary likelihood information for the modelled hubs. We denote by
% \begin{equation}
%     \mathcal{L}_i = \bigl(\mathcal{L}_i^{(1)},\dots,\mathcal{L}_i^{(K_{\mathrm{hub}})}\bigr)
% \end{equation}
% the vector of likelihood values of point $i$ with respect to the $K$ detected hubs. Each hub $k$ has its own $HLP_k$/$LLP_k$ threshold $\mathcal{T}_k$. In practice, friends-of-friends clustering assign each point to at most one hub, so we store a single hub index $k$ for each point. 
For points that do not belong to any hub the original LAAT transition probabilities \eqref{eq:old_prob} are used. 
When the current point $\vec{x}_i$ has a valid hub assignment $k$, we replace the old transition using the hub's LLPs as follows. %with likelihood based one.
% After the High Likelihood Points have been removed, a void is left behind at each hub. The remaining Low Likelihood Points are handled differently. When an ant is on an LLP, it no longer uses the 
For a neighbour $\vec{x}_j \in \mathcal{N}_r^{(i)}$ we define the likelihood difference term $\Delta^{(i,j)}_{k} = \bigl|\mathcal{L}_k^{(i)} - \mathcal{L}_k^{(j)}\bigr|$, and the likelihood-based transition probability becomes
\begin{equation}
    P(j \mid i)
    =
    \frac{
        \Delta_{k}^{(i,j)} \,
        \sfrac{\mathcal{L}_k^{(j)}}{T_k}
    }{
        \sum_{j' \in \mathcal{N}_r^{(i)}}
        \Delta_{k}^{(i,j\prime)} \,
        \sfrac{\mathcal{L}_{k}^{(j\prime)}}{T_k}
    } .\label{eq:Pnew}
\end{equation}
% where $\mathcal{L}_i^{k}$ and $\mathcal{L}_j^{k}$ are the likelihood values at $i$ and $j$ under hub $k$, respectively, and $\mathcal{T}^{k}$ is the hub likelihood threshold separating $HLP_k$ from $LLP_k$. For neighbours without likelihood information (i.e. points that do not belong to any hub), the original LAAT probability \eqref{eq:old_prob} is used instead. 
This probability favours target neighbours that both differ strongly in likelihood from $\vec{x}_i$ and lie in high-likelihood regions, thus directing ants towards sharp likelihood gradients. %and hub centres while still allowing exploration. 
This transition alone would steer the ants to surround the cavity. To allow faster exploration, we propose a double jump strategy with controlled repulsion. 
For the target $\vec{x}_j$ from $\vec{x}_i$ \eqref{eq:Pnew} we compute the Euclidean distance $\delta_{ij}$.
With a probability $P_\text{2nd}(j)=\exp(-\sfrac{\delta_{ij}}{r})$ a 2nd jump is triggered to a random point $\vec{x}_{j\prime}$ in the LLP set of model $k$, effectively expanding the neighbourhood $\mathcal{N}_r^{(i)}$ to the model shell LLP$_k$. 
A longe-range repulsion follows the 2nd jump to increase the chance to leave the hub. 
Concretely, a shell is formed centred at $\vec{x}_{j\prime}$ with potential target points $\vec{x}_m$ at a distance 
% To further reduce trapping inside dense hubs, we introduce a two-stage controlled random jump whenever an ant is located at a LLP. The first jump helps the ant cross the curved LLP shell around the hub, while the second applies a controlled repulsion away from the dense core, pushing the ant toward lower likelihood outskirts where filaments emerge.
% First, among the neighbours in $\mathcal{N}_r^{(i)}$ that belong to the $LLP_k$ region, we identify the one with the highest likelihood and compute the Euclidean distance $\delta_{ij}$ between this neighbour $j$ and the current position $i$. If $\delta_{ij} < r$ (with $r$ the LAAT neighbourhood radius), a random jump to a low-likelihood point $j' \in LLP_k$ is triggered with probability
% \begin{equation}
%     P_{\text{random}} = \exp\!\left(-\frac{\delta_{ij}}{r}\right).
% \end{equation} After this first jump, the ant is located at $j'$ with likelihood $\mathcal{L}_{\mathrm{j'}}^{k}$. A second, longer-range jump within the same hub is then proposed by sampling a candidate shell point $m$ at a distance 
$\gamma_{\min} r \le \delta_{j'm} \le \gamma_{\max} r$. %from the current shell position. 
The parameters $\gamma_{\min}$ and $\gamma_{\max}$ control the repulsion range %of the second jump performed inside a hub, specifying an acceptable radial band, 
relative to the LAAT parameter %neighbourhood 
radius $r$. %, from which the candidate point for the second jump is drawn.
The transition probability for repulsion %prefers low-likelihood points with 
% Let $\mathcal{L}_{\mathrm{m}}$ denote the likelihood at this candidate point. The second jump is accepted with probability
% \begin{equation}
    $P_{\text{rep}}%second\_random}}
    = \exp\!\left(-\sfrac{\mathcal{L}_k^{(m)}}
                        {\mathcal{L}^{(j\prime)}_{k}}\right)
$, %\end{equation}which 
favours moves towards lower-likelihood points, % shell regions 
that connect the hub to fainter structures, effectively enabling the ants to leave the hub. %This two-step mechanism prevents excessive random wandering and steers ants towards intermediate regions that link dense clusters to fainter structures.
%Once the HLP have been removed, a void is left behind at each hub. The remaining LLP are handled differently: when an ant is on an LLP, it follows likelihood-based rather than standard LAAT probabilities. The jump preference from point $p$ to neighbour $n$ is proportional to $\lvert\mathcal{L}_p-\mathcal{L}_n\rvert\frac{\mathcal{L}_n}{\mathcal{L}_{\mathrm{thr}}}$, where $\mathcal{L}_{\mathrm{thr}}$ is the hub threshold. This approach steers ants towards regions of steep likelihood gradients and hub centres, promoting balanced exploration of dense areas. To prevent trapping further, an ant landing on a hub point identifies the neighbour with the highest likelihood at a distance of $d$. The new position $p'$ then selects a candidate hub point at distance $1.5r \le d' \le 2.5r$, accepted with probability $P_{\mathrm{second}} = e^{-\mathcal{L}_c/\mathcal{L}_p}$. These controlled jumps limit random movement and guide ants through transition zones that link dense hubs to fainter structures, thereby improving filament recovery.
\section{Experiments and Discussion}
%In this section, we show how the new algorithm outperforms the old one in the recovery of a simple mock dataset where we have groundtruths (and so we can exactly see how much of the main components of the data have been recovered). Subsequently, we analyse how sensitive the results are to different parameters. We do this using a new mock dataset that is similar to a cosmic web scenario. In addition, we show the progressess on a cube of $50^3 Mpc^3/h$ that has been selected from a cosmological simulation containing around $2.8\cdot 10^5$ particles.
We compare the new algorithm with the original LAAT on a synthetic %mock 
dataset with known ground truth components, %which allows us 
to measure component recovery directly. Then we study the new parameters sensitivity using a second cosmic-web-like mock dataset and demonstrate its %test 
scalability on a $50^3\,\mathrm{Mpc}^3/h$ cosmic web N-body simulation cube containing $2.8\times10^5$ particles. 
Throughout the experiments, we fix $\beta=10$ in the MC transition probabilities %weight definition, use Power Iteration with 
tolerance $\varepsilon=0.1$ and a maximum of 100 steps in the Power Iteration. % and a maximum of $100$ iterations, $N_{min}$ to $1000$, run the 
We model each Bayesian Gaussian mixture with maximal 10 %mixture 
components, % $K_{\max}=10$, 
full covariances and a uniform % weight 
concentration prior. % of $1/K_{\max}$, and set 
We set %the 
ant-transition parameters %to 
$\gamma_{\min}=1.5$ and $\gamma_{\max}=2.5$.

\begin{figure}
    \centering
    \includegraphics[width=\linewidth, height=0.18\textheight]{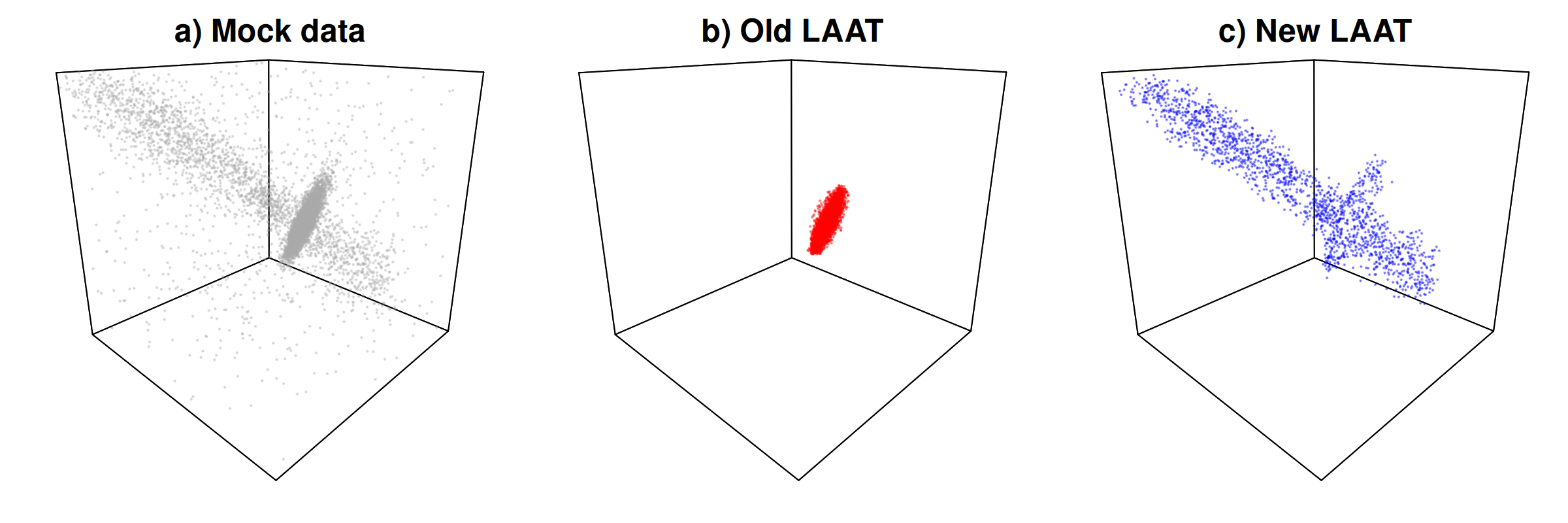}
    \caption{a) Mock data. b) and c) Old and New LAAT outputs.}
    \label{fig:simple}
\end{figure}

\begin{wraptable}[6]{r}{0.4\textwidth} 
%\vspace{-0.2cm}
  %\vspace{-\baselineskip}            
  \centering
  \caption{Component recovery.}\label{tab:recovery}
  \label{tab:simple}
  \begin{tabular}{lrr}
    \toprule
    & \textbf{OLD} & \textbf{NEW} \\
    \midrule
    Dense hub & 88\% & 5\% \\
    Filament  & 5\% & 72\% \\
    Noise     & 1\% & 3\% \\
    \bottomrule
  \end{tabular}
\end{wraptable}

\noindent\textit{Synthetic data:} the sample contains $8\times10^3$ particles: $64\%$ in the dense hub, $26\%$ in the filament, and $10\%$ noise. 
For the old method we use $100$ ants, $1000$ epochs, and $10000$ steps with $r=0.25$. 
For the new one, the new hyperparameters $(\psi,\eta,b)$ are set to $(0.005,-0.2,r)$. 
To ensure a fair comparison, we estimated, by wall-clock CPU timing, that the new preprocessing steps weight around $40$ epochs of $10000$ steps on this dataset. As a consequence, we leave the number of ants and steps unchanged, and reduce the number of epochs to $960$. 
Results are shown in %Fig. 
\autoref{fig:simple} and %Table 
\autoref{tab:simple} %, %which report 
reporting the recovery fractions of each component and demonstrate the significant improvement in filamentary component recovery achieved by the new algorithm.
%the data consists of just less than $8\cdot10^3$ particles: $64\%$ of them belong to the dense hub, $26\%$ to the filament, and the remaining $10\%$ to the noise.
%For the old method, we use $100$ ants, $1000$ epochs, $10000$ steps and set $r = 0.25$. For the new one, the new hyperparameters $(\psi, \eta, b)$ are set to $(0.005, -0.2, 0)$. To ensure a fair comparison, we estimated, by wall-clock CPU timing, that the new preprocessing steps weight around $40$ epochs of $10000$ steps on this dataset. As a consequence, we leave the number of ants and steps unchanged, and reduce the number of epochs to $960$. We show the outputs in Fig. \ref{fig:simple}, while in Table \ref{tab:simple} we report the proportion of the original component points recovered by each method.

\begin{figure}[t]
    \centering
    \includegraphics[width=\linewidth, height=0.18\textheight]{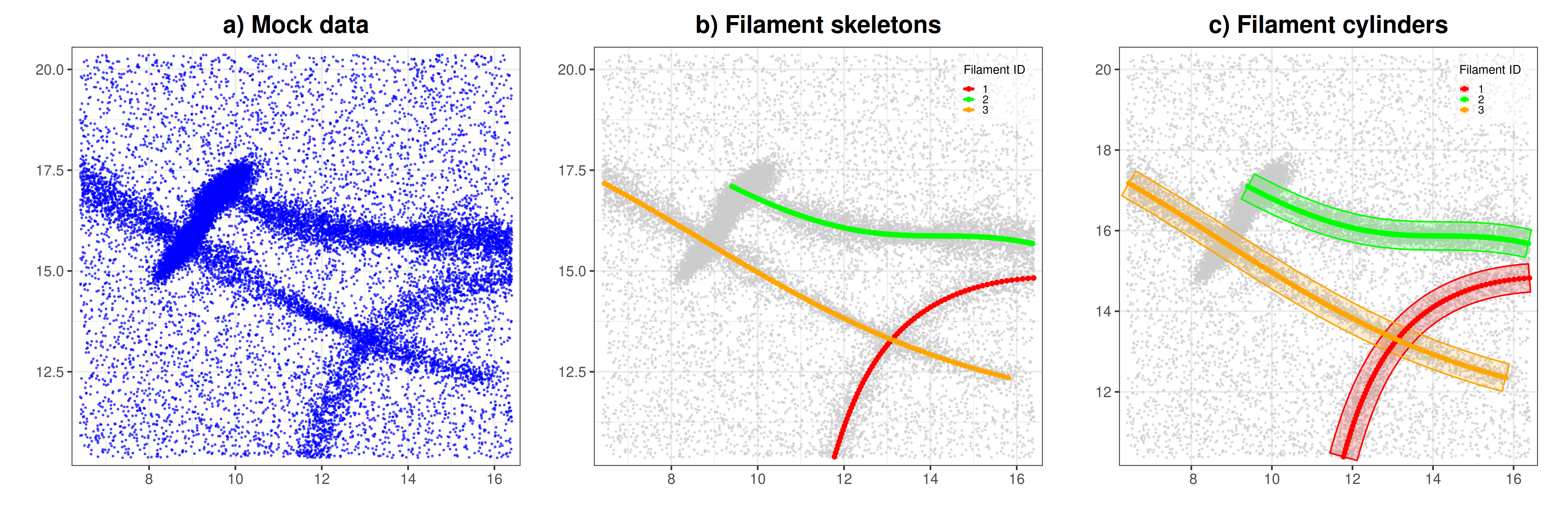} % 
    \caption{a) Mock data. b) Filament skeletons. c) %Cylindrical volumes used 
    Tubes to select candidate filament points for the analysis.}
    \label{fig:mock_cw}
\end{figure}

\noindent\textit{Parameter analysis:} %the method introduces several parameters: $\beta$ in the Markov-chain weighting, the Power Iteration tolerance and maximum iterations, a minimum hub size $N_{min}$, the BGM settings, the ant-transition parameters $(\gamma_{\min},\gamma_{\max})$, and the pre-processing parameters $(\psi,\eta,b)$.  The former parameters are fixed throughout our experiments, whereas 
the algorithm behaviour is primarily governed by the preprocessing parameters, with all others fixed due to their weaker empirical impact.
The parameter $b$ controls the typical spatial extent of detected hubs, while $\psi$ and $\eta$ regulate the partition into dense and non-dense points, balancing speed and completeness.
Lower values classify more points as dense, risking 
% which risks the 
removal of faint structures, whereas higher values make hub detection stricter and limit hub usage.
To study this trade-off, we use a synthetic, cosmic web-like point cloud comprising two high-density Gaussian nodes, three filaments of varying length, shape and density, and %a 
uniform background noise (totalling around $2.4\cdot10^4$ points).
% Following \citet{Taghribi2022}, 
We perform a sensitivity analysis of the parameters similar to \cite{Taghribi2022}. %, %by varying them and assessing the performance with a chosen metric.
Specifically, we alter $\psi\in[0.0005,0.5]$ and $\eta\in[-1,1]$, and fix the remaining parameters to $100$ ants, $100$ epochs, $5000$ steps, $r=0.5$, and $b=0.2$. 
% Figure 
\autoref{fig:mock_cw} shows the %mock 
dataset with filament skeletons and tube %cylindrical 
regions %used 
to select points at a fixed radial distance ($r=0.35$), %which corresponds roughly 
corresponding to $\sim 70\%$ of all filament points. 
As in \citet{Taghribi2022} we evaluate the recovery of the filament using the average Hausdorff distance (AHD) \cite{dubuisson}. % Unlike that work, which made comparisons with the full manifold catalogue, our
We focus %is 
specifically on filaments with comparisons made between the union of true filament points within the tubes % cylinders 
in %Figure 
\autoref{fig:mock_cw} and %the 
LAAT-selected points %, which are 
constrained to the same regions (\autoref{tab:recovery}).
\begin{wrapfigure}[13]{l}{0.48\columnwidth} 
  \centering
  \includegraphics[width=0.8\linewidth, height=0.23\textheight]{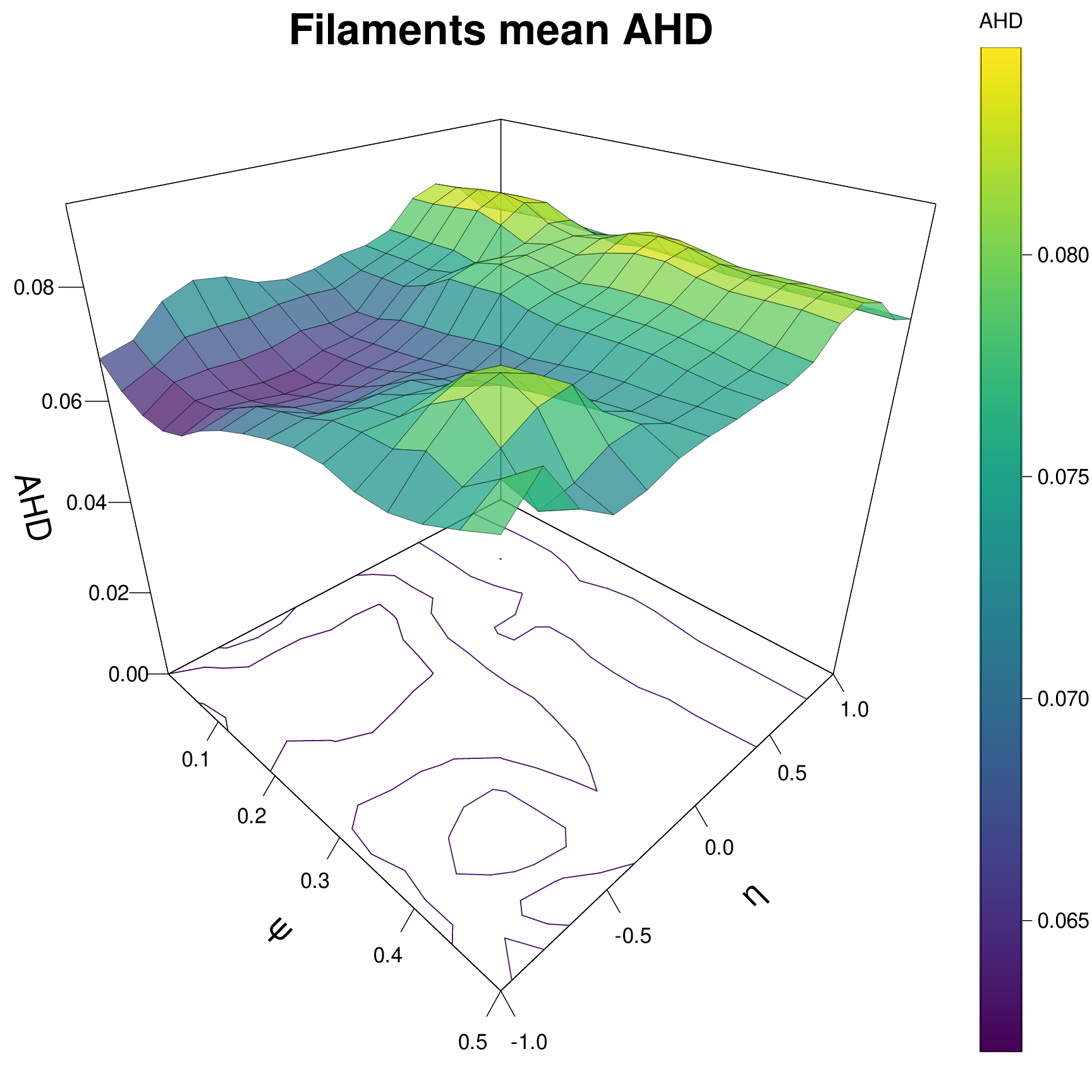}
  \caption{Global mean AHD %Average Hausdorff Distance (AHD) 
  as a function of the hyperparameters $\psi$ and $\eta$.}
  \label{fig:ahd}
\end{wrapfigure}
As shown in \autoref{fig:ahd}, increasing either $\psi$ or $\eta$ makes hub detection more restrictive, reducing filament completeness and increasing the average Hausdorff distance. % (AHD).
The best overall recovery for this dataset is obtained for $(\psi, \eta) = (0.1, -1)$, which corresponds to the lowest mean AHD.

\begin{figure}[t]
    \centering
    \includegraphics[width=\linewidth, height=0.15\textheight]{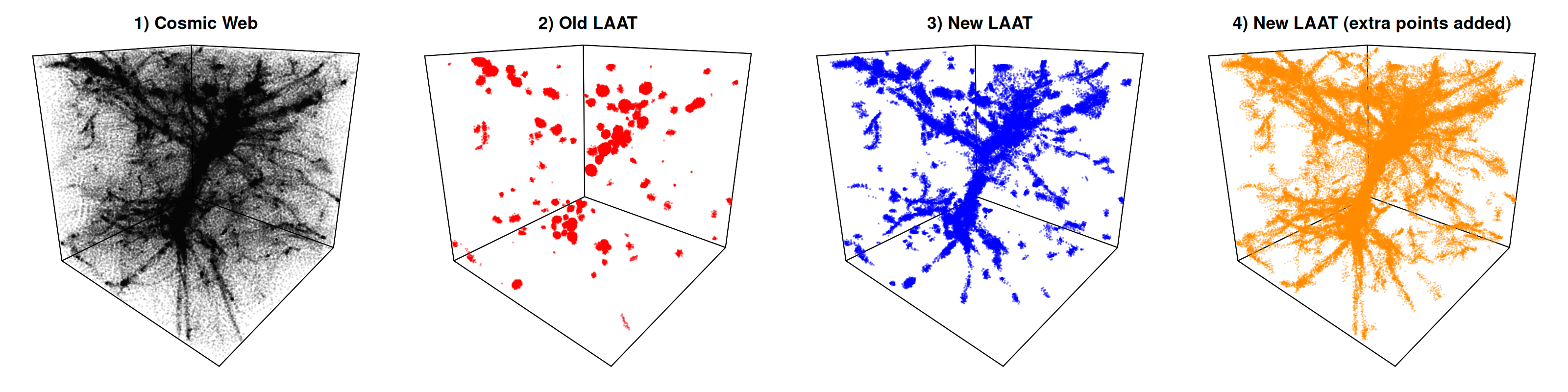}
    \caption{a) cosmic web data. b) and c) Top $%1 \cdot 
    10^5$ particles with highest pheromone levels from original and new LAAT. (d) Top $%1 \cdot 
    10^5$ particles plus %approximately 
    $\sim 3.7 \cdot 10^4$ particles (matching %equalling 
    the number %of particles %removed during 
    new LAAT preprocessing removed).} % in the new LAAT run).}
    \label{fig:cw}
\end{figure}

\textit{cosmic web:} %shows how the new version of t
The new Hub-LAAT significantly improves scenarios involving multiple dense regions, as they 
% The new algorithm can significantly improve the behaviour of LAAT in scenarios involving several dense regions. Unlike the previous implementation of LAAT, which struggled to navigate dense clusters, the new version uses them to its advantage. %It uses High-density regions %to 
guide the ants more efficiently, %thereby
accelerating and substantially improving the overall exploration process, as  
% This comparison is 
illustrated in \autoref{fig:cw}. Panels (b) and (c) show, %respectively, 
the top $10^{5}$ particles with the highest pheromone values obtained with the old and the new LAAT algorithms, using identical settings ($100$ ants, $100$ epochs, $10^{4}$ steps and $r=1$). %For the new LAAT algorithm, % run, the new 
Hub-LAAT hyperparameters $(\psi, \eta, b)$ were set to $(0.005, -0.6, r)$. 
% In the new LAAT, the stage that models the dense regions removes a fraction of the particles before pheromone deposition. 
Panel (d) %therefore 
displays the top $10^{5}$ particles from Hub-%the new 
LAAT (c) %together 
with an additional number of particles equalling those removed during preprocessing (%here 
$\sim 3.7\times 10^{4}$). Lacking ground truth, the evaluation is necessarily qualitative.
\vspace{-0.2cm}
\section{Conclusions and future work}
%\vspace{-0.2cm}
In this paper, we propose a two-stage modification to LAAT: fast hub detection with a tailored likelihood model, followed by mixed likelihood-pheromone ant transition probabilities. %exploration. 
These extensions mitigate hub-induced overconcentration, reduce computational and memory overhead, accelerate the recovery of filaments and streams, and improve robustness in noisy, high-dimensional cosmological datasets.
In future work we will develop %involve developing 
a temporal version of the methodology to model %account for 
the dynamic evolution of galaxy structures over time. 
% \vspace{-0.2\baselineskip}

\noindent\textbf{\upshape Acknowledgments}\mbox{}\\
{\footnotesize
Funded by the European Union (MSCA EDUCADO, GA 101119830). Views and opinions expressed are however those of the author(s) only and do not necessarily reflect those of the European Union. Neither the European Union nor the granting authority can be held responsible for them.
}

% ****************************************************************************
% BIBLIOGRAPHY AREA
% ****************************************************************************

\begin{footnotesize}

% IF YOU DO NOT USE BIBTEX, USE THE FOLLOWING SAMPLE SCHEME FOR THE REFERENCES
%
% ----------------------------------------------------------------------------

% IF YOU USE BIBTEX,
% - DELETE THE TEXT BETWEEN THE TWO ABOVE DASHED LINES
% - UNCOMMENT THE NEXT TWO LINES AND REPLACE 'Name_Of_Your_BibFile'

%\bibliographystyle{plainnat}
\newcommand{\nat}{Nature}
\newcommand{\mnras}{Monthly Notices of the Royal Astronomical Society}
\newcommand{\aap}{Astronomy \& Astrophysics}
\newcommand{\araa}{Annual Review of Astronomy and Astrophysics}
\newcommand{\apj}{Astrophysical Journal}
\newcommand{\aj}{The Astronomical Journal}
%\bibliography{biblio}
\printbibliography

\end{footnotesize}

% ****************************************************************************
% END OF BIBLIOGRAPHY AREA
% ****************************************************************************

\end{document}